# A Game Theoretic Approach to Class-wise Selective Rationalization

**Shiyu Chang**[1,2*]  **Yang Zhang**[1,2*]  **Mo Yu**[2*]  **Tommi S. Jaakkola**[3]
[1]MIT-IBM Watson AI Lab  [2]IBM Research  [3]CSAIL MIT
{shiyu.chang,yang.zhang2}@ibm.com  yum@us.ibm.com  tommi@csail.mit.edu

## Abstract

Selection of input features such as relevant pieces of text has become a common technique of highlighting how complex neural predictors operate. The selection can be optimized post-hoc for trained models or incorporated directly into the method itself (self-explaining). However, an overall selection does not properly capture the multi-faceted nature of useful rationales such as pros and cons for decisions. To this end, we propose a new game theoretic approach to class-dependent rationalization, where the method is specifically trained to highlight evidence supporting alternative conclusions. Each class involves three players set up competitively to find evidence for factual and counterfactual scenarios. We show theoretically in a simplified scenario how the game drives the solution towards meaningful class-dependent rationales. We evaluate the method in single- and multi-aspect sentiment classification tasks and demonstrate that the proposed method is able to identify both factual (justifying the ground truth label) and counterfactual (countering the ground truth label) rationales consistent with human rationalization. The code for our method is publicly available[2].

## 1 Introduction

Interpretability is rapidly rising alongside performance as a key operational characteristics across NLP and other applications. Perhaps the most straightforward means of highlighting how a complex method works is by selecting input features relevant for the prediction (e.g., [19]). If the selected subset is short and concise (for text), it can potentially be understood and verified against domain knowledge. The selection of features can be optimized to explain already trained models [24], incorporated directly into the method itself as in self-explaining models [19, 12], or optimized to mimic available human rationales [8].

One of the key questions motivating our work is extending how rationales are defined and estimated. The common paradigm to date is to make an *overall* selection of a feature subset that maximally explains the target output/decision. For example, maximum mutual information criterion [12, 19] chooses an overall subset of features such that the mutual information between the feature subset and the target output decision is maximized, or, equivalently, the entropy of the target output decision conditional on this subset is minimized. Rationales can be multi-faceted, however, involving support for different outcomes, just with different degrees. For example, we could understand the overall sentiment associated with a product in terms of weighing associated pros and cons contained in the review. Existing rationalization techniques strive for a single overall selection, therefore lumping together the facets supporting different outcomes.

We propose the notion of *class-wise rationales*, which is defined as multiple sets of rationales that respectively explain support for different output classes (or decisions). Unlike conventional rationalization schemes, class-wise rationalization takes a candidate outcome as input, which can be different from the ground-truth class labels, and uncovers rationales specifically for the given class. To find such rationales, we introduce a game theoretic algorithm, called *Class-wise Adversarial Rationalization* (CAR). CAR consists of three types of players: factual rationale generators, which generate rationales that are consistent with the actual label, counterfactual rationale generators, which generate rationales that counter the actual label, and discriminators, which discriminate between factual and counterfactual rationales. Both factual and counterfactual rationale generators try to competitively "convince" the discriminator that they are factual, resulting in an adversarial game between the counterfactual generators and the other two types of players.

---

[*]Authors contributed equally to this paper.
[2]https://github.com/code-terminator/classwise_rationale

We will show in a simplified scenario how CAR game drives towards meaningful class-wise rationalization, under an information-theoretic metric, which is a class-wise generalization of the maximum mutual information criterion. Moreover, empirical evaluation on both single- and multi-aspect sentiment classification show that CAR can successfully find class-wise rationales that align well with human understanding. The data and code will become publicly available.

## 2 Related Work

There are two lines of research on generating interpretable features of neural network. The first is to directly incorporate the interpretations into the models, *a.k.a* self-explaining models [3, 4, 5, 15]. The other line is to generate interpretations in a post-hoc manner. There are several ways to perform post-hoc interpretations. The first class of method is to explicitly introduce a generator that learns to select important subsets of inputs as explanations [12, 19, 21, 30, 31], which often comes with some information-theoretic properties. The second class is to evaluate the importance of each input feature via backpropagation of the prediction. Many of these methods utilize gradient information [6, 20, 25, 26, 27, 28], while techniques like local perturbations [11, 13, 16, 22] and Parzen window [7] have also been used to loose the requirement of differentiability. Finally, the third class is locally fitting a deep network with interpretable models, such as linear models [2, 24]. There are also some recent works trying to improve the fidelity of post hoc explanations by including the explanation mechanism in the training procedure [17, 18].

Although none of the aforementioned approaches can perform class-wise rationalization, gradient-based methods can be intuitively adapted for this purpose, which produces explanations toward a certain class by probing the importance with respect to the corresponding class logit. However, as noted in [24], when the input feature is far away from the corresponding class, the local gradient or perturbation probe can be very inaccurate. Evaluation of such methods will be provided in section 5.

## 3 Class-wise Rationalization

In this section, we will introduce our adversarial approach to class-wise rationalization. For notations, upper-cased letters, *e.g.* $X$ or $\boldsymbol{X}$, denote random variables or random vectors respectively; lower-cased letters, *e.g.* $x$ or $\boldsymbol{x}$, denote deterministic scalars or vectors respectively; script letters, *e.g.* $\mathcal{X}$, denote sets. $p_{X|Y}(x|y)$ denotes the probability of $X = x$ conditional on $Y = y$. $\mathbb{E}[X]$ denotes expectation.

### 3.1 Problem Formulation

Consider a text classification problem, where $\boldsymbol{X}$ is a random vector representing a string of text, and $Y \in \mathcal{Y}$ represents the class that $\boldsymbol{X}$ is in. The class-wise rationalization problem can be formulated as follows. For any input $\boldsymbol{X}$, our goal is to derive a class-wise rationale $\boldsymbol{Z}(t)$ for any $t \in \mathcal{Y}$ such that $\boldsymbol{Z}(t)$ provides evidence supporting class $t$. Each rationale can be understood as a masked version $\boldsymbol{X}$, *i.e.* $\boldsymbol{X}$ with a subset of its words masked away by a special value (*e.g.* 0). Note that class-wise rationales are defined for *every* class $t \in \mathcal{Y}$. For $t = Y$ (the correct class) the corresponding rationale is called factual; for $t \neq Y$ we call them counterfactual rationales. For simplicity, we will focus on two-class classification problems ($\mathcal{Y} = \{0, 1\}$) for the remainder of this section. Generalization to multiple classes will be discussed in appendix A.4.

As a clarification, notice that during inference, the class $t$ that is provided to the system does *not* need to be the ground truth. No matter what $t$ is provided, factual or counterfactual, the algorithm is supposed to try its best to find evidence in support of $t$. Therefore, the inference does not need to access the ground truth label. However, the training of the algorithm requires the ground truth label $Y$, because it needs to learn the phrases and sentences that are informative of each class.

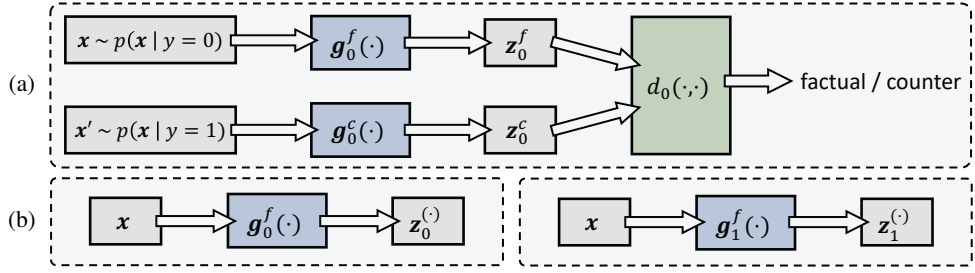

Figure 1: CAR training and inference procedures of the class-0 case. (a) The training procedure. (b) During inference, there is no ground truth label. In this case, we will always trigger the factual generators.

### 3.2 The CAR Framework

CAR uncovers class-wise rationales using adversarial learning, inspired by outlining pros and cons for decisions. Specifically, there are two *factual rationale generators*, $g_t^f(X)$, $t \in \{0, 1\}$, which generate rationales that justify class $t$ when the actual label agrees with $t$, and two *counterfactual rationale generators*, $g_t^c(X)$, $t \in \{0, 1\}$, which generate rationales for the label other than the ground truth. Finally, we introduce two discriminators $d_t(Z)$, $t \in \{0, 1\}$, which aim to discriminate between factual and counterfactual rationales, *i.e.*, between $g_t^f(X)$ and $g_t^c(X)$. We thus have six players, divided into two groups. The first group pertains to $t = 0$ and involves $g_0^f(X)$, $g_0^c(X)$ and $d_0(Z)$ as players. Both groups play a similar adversarial game, so we focus the discussion on the first group.

**Discriminator:** In our adversarial game, $d_0(\cdot)$ takes a rationale $Z$ generated by either $g_0^f(\cdot)$ or $g_0^c(\cdot)$ as input, and outputs the probability that $Z$ is generated by the factual generator $g_0^f(\cdot)$. The training target for $d_0(\cdot)$ is similar to the generative adversarial network (GAN) [14]:

$$d_0(\cdot) = \underset{d(\cdot)}{\operatorname{argmin}} -p_Y(0)\mathbb{E}[\log d(g_0^f(X))|Y = 0] - p_Y(1)\mathbb{E}[\log(1 - d(g_0^c(X)))|Y = 1]. \quad (1)$$

**Generators:** The factual generator $g_0^f(\cdot)$ is trained to generate rationales from text labeled $Y = 0$. The counterfactual generator $g_0^c(\cdot)$, in contrast, learns from text labeled $Y = 1$. Both generators try to convince the discriminator that they are factual generators for $Y = 0$.

$$g_0^f(\cdot) = \underset{g(\cdot)}{\operatorname{argmax}} \mathbb{E}[h_0(d_0(g(X)))|Y = 0], \quad \text{and} \quad g_0^c(\cdot) = \underset{g(\cdot)}{\operatorname{argmax}} \mathbb{E}[h_1(d_0(g(X)))|Y = 1],$$
$$\text{s.t.} \quad g_0^f(X)) \text{ and } g_0^c(X) \text{ satisfy some sparsity and continuity constraints.} \quad (2)$$

The constraints stipulate that the words selected as rationales should be a relatively small subset of the entire text (sparse) and they should constitute consecutive segments (continuous). We will keep the constraints abstract for generality for now. The actual form of the constraints will be specified in section 4. $h_0(\cdot)$ and $h_1(\cdot)$ are both monotonically-increasing functions that satisfy the following properties:

$$xh_0\left(\frac{x}{x+a}\right) \text{ is convex in } x, \quad \text{and} \quad xh_1\left(\frac{a}{x+a}\right) \text{ is concave in } x, \quad \forall x, a \in [0, 1]. \quad (3)$$

One valid choice is $h_0(x) = \log(x)$ and $h_1(x) = -\log(1 - x)$, which reduces the problem to the more canonical GAN-style problem. In practice, we find that other functional forms have more stable training behavior. As shown later, this generalization is closely related to $f$-divergence.

Figure 1(a) summarizes the training procedure of these three players. As can be seen, $g_0^c(\cdot)$ plays an adversarial game with both $d_0(\cdot)$ and $g_0^f(\cdot)$, because it tries to trick $d_0(\cdot)$ into misclassifying its output as factual, whereas $g_0^f(\cdot)$ helps $d_0(\cdot)$ make the correct decision. The other group of players, $g_1^f(\cdot)$, $g_1^c(\cdot)$ and $d_1(\cdot)$, play a similar game. The only difference is that now the factual generator operates on text with label $Y = 1$, and the counterfactual generator on text with label $Y = 0$.

### 3.3 How Does It Work?

Consider a simple bag-of-word scenario, where the input text is regarded as a collection of words drawn from a vocabulary of size $N$. In this case, $\boldsymbol{X}$ can be formulated as an $N$-dimensional binary vector. $\boldsymbol{X}_i = 1$, if the $i$-th word is present, and $\boldsymbol{X}_i = 0$ otherwise. $p_{\boldsymbol{X}|Y}(\boldsymbol{x}|y)$ represents the probability distribution of $\boldsymbol{X}$ in natural text conditional on different classes $Y = y$.

The rationales $\boldsymbol{Z}_0^f$ and $\boldsymbol{Z}_0^c$ are also multivariate binary vectors. $\boldsymbol{Z}_{0,i}^f = 1$ if the $i$-th word is selected as part of the factual rationale, and $\boldsymbol{Z}_{0,i}^f = 0$ otherwise. $p_{\boldsymbol{Z}_0^f|Y}(\boldsymbol{z}|0)$ denotes the *induced* distribution of the factual rationales, which is only well-defined in the factual case ($Y = 0$). This distribution is determined by how $g_0^f(\cdot)$ generates the rationales across examples. In the optimization problem, we will primarily make use of the induced distribution, and similarly for the counterfactual rationales.

To simplify our discussion, we assume that the dimensions of $\boldsymbol{X}$ are independent conditional on $Y$. Furthermore, we assume that the rationale selection scheme selects each word independently, so the induced distributions over $\boldsymbol{Z}_0^f$ and $\boldsymbol{Z}_0^c$ are also independent across dimensions, conditional on $Y$. Formally, $\forall \boldsymbol{x}, \boldsymbol{z} \in \{0,1\}^N, \forall y \in \{0,1\}$,

$$p_{\boldsymbol{X}|Y}(\boldsymbol{x}|y) = \prod_{i=1}^N p_{\boldsymbol{X}_i|Y}(\boldsymbol{x}_i|y), \quad p_{\boldsymbol{Z}_0^f|Y}(\boldsymbol{z}|y) = \prod_{i=1}^N p_{\boldsymbol{Z}_{0,i}^f|Y}(\boldsymbol{z}_i|y), \quad p_{\boldsymbol{Z}_0^c|Y}(\boldsymbol{z}|y) = \prod_{i=1}^N p_{\boldsymbol{Z}_{0,i}^c|Y}(\boldsymbol{z}_i|y). \quad (4)$$

Figure 2(left) plots $p_{\boldsymbol{X}_i|Y}(1|0)$ and $p_{\boldsymbol{X}_i|Y}(1|1)$ as functions of $i$ (the horizontal axis corresponds to sorted word identities). These two curves represent the occurrence of each word in the two classes. In the figure, the words to the left satisfy $p_{\boldsymbol{X}_i|Y}(1|0) > p_{\boldsymbol{X}_i|Y}(1|1)$, *i.e.* they occur more often in class 0 than in class 1. These words are most indicative of class 0, which we will call *class-0 words*. Similarly, the words to the right are called *class-1 words*.

Figure 2(left) also plots an example of $p_{\boldsymbol{Z}_{0,i}^f|Y}(1|0)$ and $p_{\boldsymbol{Z}_{0,i}^c|Y}(1|1)$ curves (solid, shaded curves), which represents the occurrence of each word in the factual and counterfactual rationales respectively. Note that these two curves must satisfy the following constraints:

$$p_{\boldsymbol{Z}_{0,i}^f|Y}(1|0) \leq p_{\boldsymbol{X}_i|Y}(1|0), \quad \text{and} \quad p_{\boldsymbol{Z}_{0,i}^c|Y}(1|1) \leq p_{\boldsymbol{X}_i|Y}(1|1). \quad (5)$$

This is because a word can be chosen as a rationale *only if* it appears in a text, and this strict relation translates into an inequality constraint in terms of the induced distributions. As shown in figure 2(left), the $p_{\boldsymbol{Z}_{0,i}^f|Y}(1|0)$ and $p_{\boldsymbol{Z}_{0,i}^c|Y}(1|1)$ curves are always below the $p_{\boldsymbol{X}_i|Y}(1|0)$ and $p_{\boldsymbol{X}_i|Y}(1|1)$ curves respectively. For the remainder of this section, we will refer to $p_{\boldsymbol{X}_i|Y}(1|0)$ as the *factual upper-bound*, and $p_{\boldsymbol{X}_i|Y}(1|1)$ as the *counterfactual upper-bound*. What we intend to show is that the optimal strategy for both rationale generators in this adversarial game is to choose the class-0 words.

**The optimal strategy for the counterfactual generator:** We will first find out what is the optimal strategy for the counterfactual generator, or, equivalently, the optimal $p_{\boldsymbol{Z}_{0,i}^c|Y}(1|1)$ curve, given an arbitrary $p_{\boldsymbol{Z}_{0,i}^f|Y}(1|1)$ curve. The goal of the counterfactual generator is to fool the discriminator. Therefore, its optimal strategy is to match the the counterfactual rationale distribution with the factual rationale distribution. As shown in figure 2(middle), the $p_{\boldsymbol{Z}_{0,i}^c|Y}(1|1)$ (blue) curve tries to overlay with the $p_{\boldsymbol{Z}_{0,i}^f|Y}(1|1)$ (green) curve, within the limits of the counterfactual upper bound constraint.

**The optimal strategy for the factual generator:** The goal of the factual generator is to help the discriminator. Therefore, its optimal strategy, given the optimized counterfactual generator, is to "steer" the factual rationale distribution away from the counterfactual rationale distribution. Recall that the counterfactual rationale distribution always tries to match the factual rationale distribution, unless its upper-bound is binding. The factual generator will therefore choose the words whose factual upper-bound is much higher than the counterfactual upper-bound. These words are, by definition, most indicative of class 0. The counterfactual generator will also favor the same set of words, due to its incentive to match the distributions. Figure 2(right) illustrates the optimal strategy for the factual rationale under sparsity constraint

$$\sum_{i=1}^N \mathbb{E}[\boldsymbol{Z}_{0,i}^f] = \sum_{i=1}^N p_{\boldsymbol{Z}_{0,i}^f|Y}(1|1) \leq \alpha. \quad (6)$$

The left-hand side in equation (6) represents the expected factual rationale length (in number of words). It also represents the area under the $p_{\boldsymbol{Z}_{0,i}^f|Y}(1|1)$ curve (the green shaded areas in figure 2).

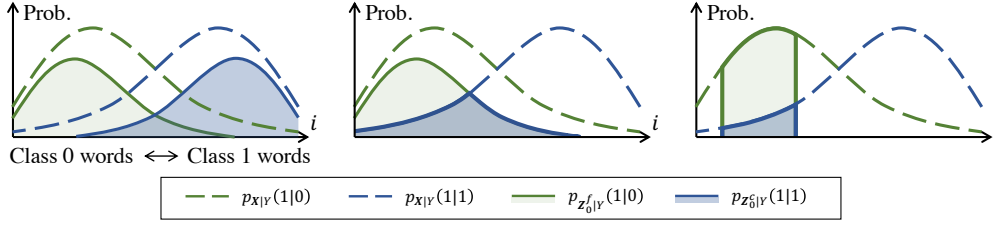

Figure 2: An illustration of how CAR works in the bag-of-word scenario with independence assumption (equation (4)). Left: example probability of occurrence of each word in the rationales from each class (solid lines), upper bounded by the probability of occurrence of each word in the natural text from each class (dashed lines). Middle: the optimal strategy for the counterfactual rationale is to match the factual rationale distribution, unless prohibited by the upper-bound. Right: the optimal strategy for the factual rationale is to steer away from the counterfactual rationale distribution, leveraging the upper-bound difference.

### 3.4 Information-theoretic Analysis

Now we are ready to embark on a more formal analysis of the effectiveness of the CAR framework, as stated in the following theorem.

**Theorem 1.** *In the bag-of-word scenario with the independence assumption as in equation* (4)*:*

**(1)** *Given the optimal $d_0(\cdot)$ and an arbitrary $\boldsymbol{g}_0^f(\cdot)$, the optimal $\boldsymbol{g}_0^c(\cdot)$ to equation* (2) *(left) will generate the counterfactual rationales that follow the following distribution:*

$$p_{\boldsymbol{Z}_{0,i}^c|Y}(1|1) = \min\left\{p_{\boldsymbol{Z}_{0,i}^f|Y}(1|0), p_{\boldsymbol{X}_i|Y}(1|1)\right\}. \tag{7}$$

**(2)** *Under some additional assumptions (see appendix A.1), given the optimal $d_0(\cdot)$ and the optimal $\boldsymbol{g}_0^c(\cdot)$, the optimal $\boldsymbol{g}_0^f(\cdot)$ to equation* (2) *(right) subject to the sparsity constraint as in equation* (6) *is given by $\boldsymbol{Z}_{0,i}^f = \boldsymbol{X}_{\mathcal{I}^*}$, where*

$$\mathcal{I}^* = \operatorname*{argmax}_{\mathcal{I}} \mathbb{E}_{\boldsymbol{X} \sim p_{\boldsymbol{X}|Y}(\cdot|0)}\left[h\left(\frac{p_{\boldsymbol{X}_{\mathcal{I}}|Y}(\boldsymbol{X}_{\mathcal{I}}|0)}{p_{\boldsymbol{X}_{\mathcal{I}}}(\boldsymbol{X}_{\mathcal{I}})}\right)\right], \quad s.t. \quad p_{\boldsymbol{X}_i|Y}(1|0) > p_{\boldsymbol{X}_i|Y}(1|1), \forall i \in \mathcal{I}, \tag{8}$$

*where $\boldsymbol{X}_{\mathcal{I}}$ denotes a subvector of $\boldsymbol{X}$ containing $\boldsymbol{X}_i, \forall i \in \mathcal{I}$.*

The proof will be given in the appendix. To better understand equation (8), it is useful to first write down the mutual information between $\boldsymbol{X}_{\mathcal{I}}$ and $Y$, a similar quantity to which has been applied to the maximum mutual information criterion [12, 19].

$$I(Y; \boldsymbol{X}_{\mathcal{I}}) = \mathbb{E}_{\boldsymbol{X},Y \sim p_{\boldsymbol{X},Y}(\cdot,\cdot)}\left[\log\left(\frac{p_{\boldsymbol{X}_{\mathcal{I}}|Y}(\boldsymbol{X}_{\mathcal{I}}|Y)}{p_{\boldsymbol{X}_{\mathcal{I}}}(\boldsymbol{X}_{\mathcal{I}})}\right)\right] = \sum_{y=0}^{1} p_Y(y) \mathbb{E}_{\boldsymbol{X} \sim p_{\boldsymbol{X}|Y}(\cdot|y)}\left[\log\left(\frac{p_{\boldsymbol{X}_{\mathcal{I}}|Y}(\boldsymbol{X}_{\mathcal{I}}|y)}{p_{\boldsymbol{X}_{\mathcal{I}}}(\boldsymbol{X}_{\mathcal{I}})}\right)\right]. \tag{9}$$

As can be seen, there is a correspondence between equations (8) and (9). First, the $\log(\cdot)$ function in equation (9) is generalized a wider selection of functional forms, $h(\cdot)$. As will be shown in the appendix A.2, equation (8) applies the $f$-divergence [1], which is a generalization to the KL-divergence as applied in equation (9). Second, notice that equation (9) is decomposed into two class-dependent terms, while equation (8) is for class-0 generators only. It can be easily shown that the class-1 generators come with a similar theoretical guarantee that corresponds to the term with $y = 1$. Therefore, the target function in equation (8) can be considered as the component in the mutual information that is specifically related to class $0$. Hence we call it *class-wise mutual information*.

### 3.5 Coping with Degeneration

It has been pointed out in [32] that the existing generator-predictor framework in [12] and [19] can suffer from the problem of *degeneration*. Since the generator-predictor framework aims to maximize the predictive accuracy of the predictor, the generator and predictor can collude by selecting uninformative symbols to encode the class information, instead of selecting words and phrases that truly explain the class. For example, consider the following punctuation communication scheme: when $Y = 0$, the rationale would select only one comma ","; when $Y = 1$, the rationale would select only one period ".". This rationalization scheme guarantees a high predictive accuracy. However, this is apparently not what we expect. Such cases are called degeneration.

From section 3.3, we can conclude that CAR will not suffer from degeneration. This is because if the factual rationale generators attempt to select uninformative words or symbols like punctuation (*i.e.* words in the middle of the $x$-axis in figure 2), then the factual rationale distribution can be easily matched by the counterfactual rationale distribution. Therefore, this strategy is not optimal for the factual generators, whose goal is to avoid being matched by the counterfactual generators.

# 4 Architecture Design and Implementation

**Architecture with parameter sharing:** In our actual implementation, we impose parameter sharing among the players. This is motivated by our observation in sections 3.3 and 3.4 that both the factual and counterfactual generators adopt the same rationalization strategy upon reaching the equilibrium. Therefore, instead of having two separate networks for the two generators, we introduce one unified generator network for each class, a class-0 generator and a class-1 generator, with the ground truth label $Y$ as an additional input to identify between factual and counterfactual modes. Specifically, $\boldsymbol{g}_0^c(\cdot)$ and $\boldsymbol{g}_0^f(\cdot)$ now share the same parameters in a single generator network $\boldsymbol{g}_0(\cdot, Y)$, where $\boldsymbol{g}_0^f(\cdot) = \boldsymbol{g}_0(\cdot, 0)$, and $\boldsymbol{g}_0^c(\cdot) = \boldsymbol{g}_0(\cdot, 1)$. Please note that after the parameter sharing, $\boldsymbol{g}_0(\cdot, 0)$ and $\boldsymbol{g}_0(\cdot, 1)$ are still considered as two distinct players, in the sense that they are still trained to optimize different target functions (equation (2)), and they still play the same adversarial game with each other. Similarly, $\boldsymbol{g}_1^c(\cdot)$ and $\boldsymbol{g}_1^f(\cdot)$ share the same parameters in a single generator network $\boldsymbol{g}_1(\cdot, Y)$. We also impose parameter sharing between the two discriminators, $d_0(\cdot)$ and $d_1(\cdot)$, by introducing a unified discriminator, $d(\cdot, t)$, with an additional input $t$ to identify between the class-0 and class-1 cases. The trainable parameters are significantly reduced with parameter sharing.

Both the generators and the discriminators consist of a word embedding layer, a bi-direction LSTM layer followed by a linear projection layer. The generators generate the rationales by the independent selection process as proposed in [19]. At each word position $k$, the convolutional layer outputs a quantized binary mask $\boldsymbol{S}_k$, which equals to 1 if the $k$-th word is selected and 0 otherwise. The binary masks are multiplied with the corresponding words to produce the rationales. For the discriminators, the outputs of all the times are max-pooled to produce the factual/counterfactual decision.

For parameter sharing, we append the input class as a one-hot vector to each word embedding vector in both the generators and the discriminator. For the generators, the groundtruth class label $Y$ of each instance is appended; while for the discriminator, the class of generator $t$ used for generating the input rationale is appended.

**Training:** The training objectives are essentially equations (1) and (2). The only difference is that we instantiate the constraints in equation (2) transform it into a multiplier form. Specifically, the multiplier terms (or the regularization terms) are

$$\lambda_1 \left| \frac{1}{K} \mathbb{E}[\|\boldsymbol{S}\|_1] - \alpha \right| + \lambda_2 \mathbb{E}\left[ \sum_{t=2}^{K} |\boldsymbol{S}_k - \boldsymbol{S}_{k-1}| \right], \tag{10}$$

where $K$ denotes the number of words in the input text. The first term constrains on the sparsity of the rationale. It encourages that the percentage of the words being selected as rationales is close to a preset level $\alpha$. The second term constrains on the continuity of the rationale. $\lambda_1$, $\lambda_2$ and $\alpha$ are hyperparameters. The constraint is slightly different from the one in [19] in order have a more precise control of the sparsity level. The $h_0(\cdot)$ and $h_1(\cdot)$ functions in equation (2) are set to $h_0(\boldsymbol{x}) = h_1(\boldsymbol{x}) = \boldsymbol{x}$, which empirically shows good convergence performance, and which can be shown to satisfy equation (3). To resolve the non-differentiable quantization operation that produces $\boldsymbol{S}_t$, we apply the straight-through gradient computation technique [9]. The training scheme involves the following alternate stochastic gradient descent. First, the class-0 generator and the discriminator are updated jointly by passing one batch of data into the class-0 generator, and the resulting rationales, which contain both factual and counterfatual rationales depending on the actual class, are fed into the discriminator with $t = 0$. Then, the class-1 generator and the discriminator are updated jointly in a similar fashion with $t = 1$.

**Inference:** During the inference, the ground truth label is unavailable for fair comparisons with the baselines, therefore we have no oracle knowledge of which class is factual and which is counterfactual. In this case, we always trigger the factual generators, no matter what the ground truth is, as shown in figure 1(b). This is again justified by our observation in sections 3.3 and 3.4 that both the factual and counterfactual modes adopt the same rationalization strategy upon reaching the equilibrium. The only reason why we favor the factual mode to the counterfactual mode is that the former has more exposure to the words it is supposed to select during training.

## 5 Experiments

### 5.1 Datasets

To evaluate both factual and counterfactual rationale generation, we consider the following three binary classification datasets. The first one is the single-aspect Amazon reviews [10] (book and electronic domains), where the input texts often contain evidence for both positive and negative sentiments. We use predefined rules to parse reviews containing comments on both the pros and cons of a product, which is further used for automatic evaluations. We also evaluate algorithms on the multi-aspect beer [23] and hotel reviews [29] that are commonly used in the field of rationalization [8, 19]. The labels of the beer review dataset are binarized, resulting in a harder rationalization task than in [19]. The multi-aspect review is considered as a more challenging task, where each review contains comments on different aspects. However, unlike the Amazon dataset, both beer and hotel datasets only contain factual annotations. The construction of evaluation tasks is detailed in appendix B.1.

### 5.2 Baselines

**RNP:** A generator-predictor framework proposed by Lei *et al.* [19] for rationalizing neural prediction (RNP). The generator selects text spans as rationales which are then fed to the predictor for label classification. The selection maximizes the predictive accuracy of the target output and is constrained to be sparse and continuous. RNP is only able to generate factual rationales.

**POST-EXP:** The post-explanation method generates rationales of both positive and negative classes based on a pre-trained predictor. Given the predictor trained on full-text inputs, we train two separate generators $g_0(X)$ and $g_1(X)$ on the data to be explained. $g_0(X)$ always generate rationales for the negative class and $g_1(X)$ always generate rationales for the positive class. The two generators are trained to maximize the respective logits of the fixed predictor subject to sparsity and continuity regularizations, which is closely related to gradient-based explanations [20].

To seek fair comparisons, the predictors of both RNP and POST-EXP and the discriminator of CAR are of the same architecture; the rationale generators in all three methods are of the same architecture. The hidden state size of all LSTMs is set to 100. In addition, the sparsity and continuity constraints are also in the same form as our method. It is important pointing out that CAR does not use any ground truth label for generating rationales, which follows the procedures discussed in section 4.

### 5.3 Experiment Settings

**Objective evaluation:** We compare the generated rationales with the human annotations and report the precision, recall and F1 score. To be consistent with previous studies [19], we evaluate different algorithms conditioned on a similar *actual* sparsity level in factual rationales. Specifically, the target factual sparsity level is set to around ($\pm 2\%$) 20% for the Amazon dataset and 10% for both beer and hotel review. The reported performances are based on the best performance of a set of hyperparameter values. For details of the setting, please refer to appendix B.2.

**Subjective evaluation:** We also conduct subjective evaluations via *Amazon Mechanical Turk*. Specifically, we reserve 100 randomly balanced examples from each dev set for the subjective evaluations. For the single-aspect dataset, the subject is presented with either the factual rationale or the counterfactual rationale of a text generated by one of the three methods (unselected words blocked). For the factual rationales, a success is credited when the subject correctly guess the ground-truth sentiment; for the counterfactual rationales, a success is credited when the subject is convinced to choose the opposite sentiment to the ground-truth. For the multi-aspect datasets, we introduce a much harder test. In addition to guessing the sentiment, the subject is also asked to guess what aspect the rationale is about. A success is credited only when both the intended sentiment *and* the correct aspect are chosen. Under this criterion, a generator that picks the sentiment words only will score poorly. We then compute the success rate as the performance metric. The test cases are randomly shuffled.

Table 1: Objective performances of selected rationales of the Amazon review dataset. The numbers in each column represent the sparsity level, precision, recall, and F1 score, respectively. Each domain is trained independently. All results are calculated in a "micro" perspective.

| Amazon | Book | | Electronic | |
|---|---|---|---|---|
| | Factual | Counterfactual | Factual | Counterfactual |
| RNP [19] | 18.6/55.1/20.1/29.5 | - | 20.7/49.7/22.8/31.3 | - |
| POST-EXP | 20.2/64.5/28.8/39.8 | 27.9/70.2/**35.8**/**47.4** | 18.6/64.1/27.8/38.8 | 15.3/72.6/**19.5**/**30.7** |
| CAR | 20.9/**68.7**/**31.9**/**43.6** | 15.2/**72.2**/20.2/31.5 | 21.2/**70.0**/**34.7**/**46.4** | 10.2/**76.4**/13.6/23.1 |

Table 2: Objective performances of selected factual rationales for both (a) beer and (b) hotel review datasets. Each aspect is trained independently. S, P, R, and F1 indicate the sparsity level, precision, recall, and F1 score.

(a)

| Beer | Appearance | | | | Aroma | | | | Palate | | | |
|---|---|---|---|---|---|---|---|---|---|---|---|---|
| | S | P | R | F1 | S | P | R | F1 | S | P | R | F1 |
| RNP [19] | 11.9 | 72.0 | 46.1 | 56.2 | 10.7 | **70.5** | **48.3** | **57.3** | 10.0 | 53.1 | 42.8 | 47.5 |
| POST-EXP | 11.9 | 64.2 | 41.4 | 50.4 | 10.3 | 50.0 | 33.1 | 39.8 | 10.0 | 33.0 | 26.5 | 29.4 |
| CAR | 11.9 | **76.2** | **49.3** | **59.9** | 10.3 | 50.3 | 33.3 | 40.1 | 10.2 | **56.6** | **46.2** | **50.9** |

(b)

| Hotel | Location | | | | Service | | | | Cleanliness | | | |
|---|---|---|---|---|---|---|---|---|---|---|---|---|
| | S | P | R | F1 | S | P | R | F1 | S | P | R | F1 |
| RNP [19] | 10.9 | 43.3 | 55.5 | 48.6 | 11.0 | 40.0 | 38.2 | 39.1 | 10.6 | 30.5 | **36.0** | 33.0 |
| POST-EXP | 8.9 | 30.4 | 31.8 | 31.1 | 10.0 | 32.5 | 28.3 | 30.3 | 9.2 | 23.0 | 23.7 | 23.3 |
| CAR | 10.6 | **46.6** | **58.1** | **51.7** | 11.7 | **40.7** | **41.4** | **41.1** | 9.9 | **32.3** | 35.7 | **33.9** |

The subjects have to meet certain English proficiency and are reminded that some of the generated rationales are intended to trick them via word selections and masking (*e.g.* masking the negation words). Appendix B.2 contains a screenshot and the details of the online evaluation setups.

### 5.4 Results

Table 1 shows the objection evaluation results for both factual and counterfactual rationales on Amazon reviews. Constrained to highlighting 20% of the inputs, CAR consistently surpasses the other two baselines in the factual case for both domains. Compared to the POST-EXP, our method generates the counterfactual rationales with higher precision. However, since the sparsity constraint regularizes both factual and counterfactual generations and the model selection is conducted on factual sparsity only, we cannot control counterfactual sparsity among different algorithms. POST-EXP tends to highlight much more text, resulting in higher recall and F1 score. However, as will be seen later, the human evaluators still favor the counterfactual rationales generated by our algorithm.

Since the beer and hotel datasets contain factual annotations only, we report objective evaluation results for the factual rationales in table 2. CAR achieves the best performances in five out of the six cases in the multi-aspect setting. Specifically, for the hotel review, CAR achieves the best performance almost in all three aspects. Similarly, CAR delivers the best performance for the appearance and palate aspects of the beer review dataset, but fails on the aroma aspect. One possible reason for the failure is that compared to the other aspects, the aroma reviews often have annotated ground truth containing mixed sentiments. Therefore, CAR has low recalls of these annotated ground truth even when it successfully selects all the correct class-wise rationales. Also to fulfill the sparsity constraint, sometimes CAR has to select irrelevant aspect words with the desired sentiment, which decreases the precision. Illustrative examples of the described case can be found in appendix B.3. Please note that the RNP is not directly comparable to the results in [19], because the labels are binarized under our experiment setting.

We visualize the generated rationales on the appearance aspect of beer reviews in figure 3. More examples of other datasets can be found in appendix B.3. We observe that the CAR model is able to produce meaningful justifications for both factual and counterfactual labels. The factual generator picks "*two inches of frothy light brown head with excellent retention*" while the counterfactual one picks "*really light body like water*". By reading these selected texts alone, humans will easily predict a positive sentiment for the first case and be tricked for the counterfactual case.

At last, we present the subjective evaluations in figure 4. Similar to the observations in the objective studies, CAR achieves the best performances in almost all cases with two exceptions. The first one is

*Beer - Appearance*  Label - Positive

poured into pint glass . a : used motor **oil color . two inches of frothy light brown head with excellent retention and quite a bit** of lacing . nice cascade going for a while . s : oatmeal is the biggest component of the aroma . not any hops content . a bit fusely and a bit of alcohol . t : tastes like slightly sour nothing . i do n't know **what the** hell made this dark because their is no crystal malt or roasted barley component in the taste . this sucks . m : **light body , really light body like water . carbonation** is fine , but that 's about it . d : this is slightly sour water . how does anybody like this ?

Figure 3: Examples of CAR generated rationales on the appearance aspect of the beer reviews. All selected words are **bold and underlined**. Factual generation uses **blue highlight** while the counterfactual uses **red one**.

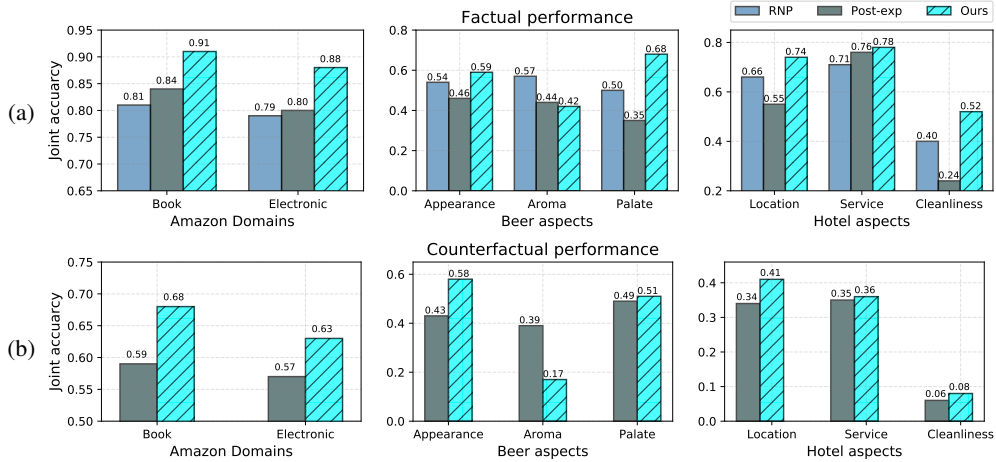

Figure 4: Subjective performances of generated rationales for both (a) factual and (b) counterfactual cases. For the Amazon reviews, subjects are asked to guess the sentiment based on the generated rationales, which random guess will have $50\%$ accuracy. For multi-aspect beer and hotel reviews, subjects need to guess both the sentiment and what aspect the rationale is about, which makes random guess only $16.67\%$.

the aroma aspect of the beer reviews, of which we have discussed the potential causes already. The second one is the counterfactual performance on the cleanliness aspect of the hotel reviews, where both POST-EXP and CAR fail to trick human. One potential reason is that the reviews on cleanliness is often very short and the valence is very clear without a mix of sentiments. Thus, it is very challenging to generate counterfactual rationales to trick a human. This can be verified by the analysis in appendix B.3. Specifically, according to figure 4, $69\%$ of the time CAR is able to trick people to guess the counterfactual sentiment, but often with the rationales extracted from the other aspects.

## 6  Conclusion

In this paper, we propose a game theoretic approach to class-wise rationalization, where the method is trained to generate supporting evidence for any given label. The framework consists of three types of players, which competitively select text spans for both factual and counterfactual scenarios. We theoretically demonstrate the proposed game theoretic framework drives the solution towards meaningful rationalizations in a simplified case. Extensive objective and subjective evaluations on both single- and multi-aspect sentiment classification datasets demonstrate that CAR performs favorably against existing algorithms in terms of both factual and counterfactual rationale generations.

## Acknowledgment

We would like to thank Yujia Bao, Yujie Qian, and Jiang Guo from the MIT NLP group for their insightful discussions. We also want to thank Prof. Regina Barzilay for her support and help.

## A Further Theoretical Discussions and Proofs

In this section, we further our discussion in section 3.4.

### A.1 Proof of Theorem 1

We will now formally prove theorem 1. First, we will briefly state the following lemma regarding the discriminator's optimal strategy.

**Lemma 1.1.** *The global optimizer of Eq. (1) is given by*

$$d_0(\boldsymbol{z}) = \frac{p_{\boldsymbol{Z}_0^f, Y}(\boldsymbol{z}, 0)}{p_{\boldsymbol{Z}_0^f, Y}(\boldsymbol{z}, 0) + p_{\boldsymbol{Z}_0^c, Y}(\boldsymbol{z}, 1)}. \tag{11}$$

The proof is similar to [14] and is omitted here.

Theorem 1 is divided into two parts ((1) and (2)). Here we will separately restate and prove each.

**Theorem 1.1. (Restating theorem 1 part (1))** *Assuming equation (4), and given that equations (11) is satisfied, the optimal solution to equation (2)(left) is given by equation (7).*

*Proof.* For notational ease, we denote

$$\begin{aligned} p_{\boldsymbol{Z}_{0,i}^c|Y}(\boldsymbol{z}_i|1) = q_i(\boldsymbol{z}_i), \quad p_{\boldsymbol{Z}_{0,i}^f|Y}(\boldsymbol{z}_i|0) = p_i(\boldsymbol{z}_i), \\ p_{\boldsymbol{X}_i|Y}(\boldsymbol{x}_i|1) = Q_i(\boldsymbol{x}_i), \quad p_{\boldsymbol{X}_i|Y}(\boldsymbol{x}_i|0) = P_i(\boldsymbol{x}_i). \end{aligned} \tag{12}$$

This notation will be used throughout the proofs in this appendix.

Under equations (4) (11), the optimization problem in equation (2)(left) can be rewritten as

$$\max_{\{q_i(1)\}} \sum_{\boldsymbol{z} \in \{0,1\}^N} \prod_i q_i(\boldsymbol{z}_i) h_1 \left( \frac{p_Y(0) \prod_i p_i(\boldsymbol{z}_i)}{p_Y(0) \prod_i p_i(\boldsymbol{z}_i) + p_Y(1) \prod_i q_i(\boldsymbol{z}_i)} \right), \\ \text{s.t.} \quad q_i(1) = 1 - q_i(0) \quad \forall i \\ 0 \leq q_i(1) \leq Q_i(1) \quad \forall i. \tag{13}$$

For each integer $j < N$, it can be easily shown that each summand in equation (13) is concave in $q_j(1)$ (from equation (3)), hence the summation is concave. Also, taking the derivative w.r.t $q_j(1)$ yields

$$\sum_{\boldsymbol{z}_{-j} \in \{0,1\}^{N-1}} p_Y(0) \prod_{i \neq j} q_i(\boldsymbol{z}_i) \{[h_1(\rho_1(\boldsymbol{z}_{-j})) - h_1'(\rho_1(\boldsymbol{z}_{-j}))\rho_1(\boldsymbol{z}_{-j})(1 - \rho_1(\boldsymbol{z}_{-j}))] \\ - [h_1(\rho_0(\boldsymbol{z}_{-j})) - h_1'(\rho_0(\boldsymbol{z}_{-j}))\rho_0(\boldsymbol{z}_{-j})(1 - \rho_0(\boldsymbol{z}_{-j}))]\}. \tag{14}$$

where $\boldsymbol{z}_{-j}$ denote a subvector of $\boldsymbol{z}$ without the $j$-th element, and

$$\rho_1(\boldsymbol{z}_{-j}) = \frac{p_Y(0) p_j(1) \prod_{i \neq j} p_i(\boldsymbol{z}_i)}{p_Y(0) p_j(1) \prod_{i \neq j} p_i(\boldsymbol{z}_i) + p_Y(1) q_j(1) \prod_{i \neq j} q_i(\boldsymbol{z}_i)} \\ \rho_0(\boldsymbol{z}_{-j}) = \frac{p_Y(0)(1 - p_j(1)) \prod_{i \neq j} p_i(\boldsymbol{z}_i)}{p_Y(0)(1 - p_j(1)) \prod_{i \neq j} p_i(\boldsymbol{z}_i) + p_Y(1)(1 - q_j(1)) \prod_{i \neq j} q_i(\boldsymbol{z}_i)}. \tag{15}$$

When $q_j(1) = p_j(1)$, we have $\rho_1(\cdot) = \rho_0(\cdot)$, and the derivative is 0. Therefore the constrained maximum is achieved at $\min\{p_j(1), Q_j(1)\}$.

□

Before we prove theorem 1(b), we will study the optimal policy of the factual generator *without* the sparsity constraint (equation (6)), as stated below

**Lemma 1.2. (Optimal factual generation without the sparsity constraint)** *Assuming equation (4), and given that equations (11) and (7) are satisfied, the optimal solution to equation (2)(right) is given by*

$$p_{\boldsymbol{Z}_{0,i}^f|Y}(1|0) = \begin{cases} p_{\boldsymbol{X}_i|Y}(1|0), & \text{if } p_{\boldsymbol{X}_i|Y}(1|0) > p_{\boldsymbol{X}_i|Y}(1|1) \\ \text{anything}, & \text{otherwise} \end{cases}. \tag{16}$$

*Proof.* Under equations (11), the optimization problem in equation (2)(right) can be rewritten as

$$\max_{\{p_i(1)\}} \sum_{\boldsymbol{z}\in\{0,1\}^N} \prod_i p_i(\boldsymbol{z}_i) h_0\left(\frac{p_Y(0)\prod_i p_i(\boldsymbol{z}_i)}{p_Y(0)\prod_i p_i(\boldsymbol{z}_i) + p_Y(1)\prod_i q_i(\boldsymbol{z}_i)}\right),$$
$$\text{s.t.} \quad p_i(1) = 1 - p_i(0) \quad \forall i \qquad (17)$$
$$0 \leq p_i(1) \leq P_i(1) \quad \forall i$$
$$q_i(1) = \min\{p_i(1), Q_i(1)\} \quad \forall i.$$

It can be easily shown that, after substituting $q_i(1)$ with $\min\{p_i(1), Q_i(1)\}$, $\forall i$, the target function is constant in $p_j(1)$ when $p_j(1) \leq Q_j(1)$. When $p_j(1) > Q_j(1)$, the derivative w.r.t. $p_j(1)$ is given by

$$\sum_{\boldsymbol{z}_{-j}\in\{0,1\}^{N-1}} p_Y(0) \prod_{i\neq j} p_i(\boldsymbol{z}_i) \{[h_0(\rho_1(\boldsymbol{z}_{-j})) + h_0'(\rho_1(\boldsymbol{z}_{-j}))\rho_1(\boldsymbol{z}_{-j})(1-\rho_1(\boldsymbol{z}_{-j}))] \quad (18)$$
$$- [h_0(\rho_0(\boldsymbol{z}_{-j})) + h_0'(\rho_0(\boldsymbol{z}_{-j}))\rho_0(\boldsymbol{z}_{-j})(1-\rho_0(\boldsymbol{z}_{-j}))]\},$$

where

$$\rho_1(\boldsymbol{z}_{-j}) = \frac{p_Y(0)p_j(1)\prod_{i\neq j} p_i(\boldsymbol{z}_i)}{p_Y(0)p_j(1)\prod_{i\neq j} p_i(\boldsymbol{z}_i) + p_Y(1)Q_j(1)\prod_{i\neq j} Q_i(\boldsymbol{z}_i)} \quad (19)$$
$$\rho_0(\boldsymbol{z}_{-j}) = \frac{p_Y(0)(1-p_j(1))\prod_{i\neq j} p_i(\boldsymbol{z}_i)}{p_Y(0)(1-p_j(1))\prod_{i\neq j} p_i(\boldsymbol{z}_i) + p_Y(1)(1-Q_j(1))\prod_{i\neq j} Q_i(\boldsymbol{z}_i)}.$$

When $p_j(1) = Q_j(1)$, the derivative is 0. Considering the function is convex in $p_j(1)$, it will be monotonically increasing when $p_j(1) > Q_j(1)$.

Therefore, when $P_j(1) \leq Q_j(1)$, $p_j(1)$ is indifferent within the constraint of $[0, P_j(1)]$; when $P_j(1) > Q_j(1)$, $p_j(1)$ achieves the maximum at $P_j(1)$. □

Now we can turn our discussion back to the case where the sparsity constraint in equation (6) is imposed. First, it is very easy to notice that when the sparsity constraint is mild, the generator can always drop the dimensions where $p_{\boldsymbol{Z}_{0,i}^f|Y}(1|0) \leq p_{\boldsymbol{Z}_{0,i}^f|Y}(1|1)$, which we call insignificant dimensions for now, because their probability can be set to anything without changing the target function values. Therefore, the real nontrivial case is when $\alpha$ is so small that dropping all the insignificant dimensions would not suffice, *i.e.*

$$\alpha < \sum_i p_{\boldsymbol{X}_i|Y}(1|0)\mathbb{1}[p_{\boldsymbol{X}_i|Y}(1|0) > p_{\boldsymbol{X}_i|Y}(1|1)], \quad (20)$$

where $\mathbb{1}[\cdot]$ is the indicator function.

We will make a stronger assumption on $h_0(\cdot)$ in addtion to equation (3): the target in equation (17) or equivalently in eqaution (2)(left) under the constraints in equation (17) is jointly convex in $\{p_i(1), \forall i\}$. It can be shown that $\log(\cdot)$ as used in canonical GAN and the linear function as used by CAR both satisfy this assumption.

Now we are ready to restate and proof theorem 1 part (2).

**Theorem 1.2. (Restating theorem 1 part (2))** *Assuming equation* (4), *the optimal solution to Eq.* (17) *with the constraint in Eqs.* (6) *and* (20) *takes the following form:*

$$p_{Z_{0i|Y}^{(f)}}(1|0) = \begin{cases} p_{X_i|Y}(1|0) & \text{if } i \in \mathcal{I}^* \\ 0 & \text{if } i \in \mathcal{I}^{*c}\setminus\mathcal{J} \end{cases}, \quad (21)$$

*where $\setminus$ denotes set subtraction; $\mathcal{J}$ can contain either one element or zero. When $\mathcal{J}$ contains zero element, $\mathcal{I}^*$ satisfies equation* (8).

*Proof.* Since equation (2)(right) is monotonically increasing functions of $p_i(1)$, $\forall i$, the constraint in equation (6) is always binding. Since Eq. (2)(right) is concave, the maximization problem always yields a corner solution, *i.e.* all but at most one $p_i(1)$'s hit the lower bound 0 or upper bound $P_i(1)$. This is because where there are two $p_i(1)$'s that do not hit either bound, the target function is jointly convex with regard to these two quantities along the binding sparsity constraint line (equation (6) with equality). Moving these two quantities along the binding sparsity constraint line will further increase the target.

Notice that equation (8) is essentially the same as the optimization problem in equation (17), but with the constraint $p_i(1) \in [0, P_i(1)], \forall i$ replaced with $p_i(1) \in \{0, P_i(1)\}, \forall i$. This is because the target function in equations (8) and (17) are the same. The only difference is that in equation (8), the variable to optimize over is the index set $\mathcal{I}$. When an index $i$ is selected in $\mathcal{I}$, this is equivalent to setting $p_i(1) = P_i(1)$; when $i$ is not selected, this is equivalent to setting $p_i(1) = 0$.

Since $\{0, P_i(1)\}$ is a subset of $[0, P_i(1)]$, the maximum value in equation (8) will be no greater than the maximum value in equation (17).

If optimal $p_i(1)$s in equation (17) either hit the lower bound 0 or upper bound $P_i(1)$ (which means $\mathcal{J}$ contains no element), it will be a feasible solution to equation (8), and therefore should also be the optimal solution in equation (8). □

### A.2 Reformulating into $f$-Divergence

It turns out that equation (8) can be well interpreted using $f$-divergence. Define

$$f(x) = xh_0(x) - h_0(1). \tag{22}$$

Then the target function of equation in (8) can be rewritten as equation

$$\mathbb{E}_{\boldsymbol{X} \sim p_{\boldsymbol{X}}(\cdot)} \left[ f\left( \frac{p_{\boldsymbol{X}_{\mathcal{I}}|Y}(\boldsymbol{Z}|0)}{p_{\boldsymbol{X}_{\mathcal{I}}}(\boldsymbol{Z})} \right) \right]. \tag{23}$$

It can be easily shown that when $h_0(x) = \log(x)$ (GAN setting) and $h_0(x) = x$ (CAR setting), $f(x)$ is convex, which satisfies the definition of $f$-divergence.

Therefore, under our toy setting, our proposed rationale generator will pick words that satisfy the following two conditions:

- maximizes the $f$-divergence between $p_{\boldsymbol{X}_{\mathcal{I}}|Y}(\boldsymbol{Z}|0)$ and $p_{\boldsymbol{X}_{\mathcal{I}}}(\boldsymbol{Z})$.
- $p_{\boldsymbol{X}_i|Y}(1|0) > p_{\boldsymbol{X}_i|Y}(1|1)$, *i.e.* occurs more frequent in the factual cases than in the counterfactual cases.

### A.3 Inference with Target Label

In section 4, we have discussed how to use CAR to generate rationales without the target label. In fact, CAR can be applied for rationalization when the prediction label is available. For example, when explaining a black-box model [24], we can regard the black-box prediction as the label. In this case, we can make use of the label $Y$ to select between factual and counterfactual generators. For example, when $Y = 0$, we can use $\boldsymbol{g}_0^f(\cdot)$ to generate class-0 rationale and $\boldsymbol{g}_1^c(\cdot)$ to generate class-1 rationale.

### A.4 Generalization to Multiple Classes

So far we have limit our dicussions to two-class classification problems, but CAR can be easily generalized to multiple classes, *i.e.* $Y \in \mathcal{Y}$ where $\mathcal{Y} = \{1, \cdots, C\}$ and $C$ is any positive integer. In this case, there are $C$ factual generators, $\boldsymbol{g}_t^f(\boldsymbol{X})$, $t \in \mathcal{Y}$, each explaining towards a specific class $t$ when $Y = t$. There are $C$ counterfactual generators, $\boldsymbol{g}_t^c(\boldsymbol{X})$, $t \in \mathcal{Y}$, each explaining towards a specific class $t$ when $Y \neq t$.[3]

There are $C$ discriminators, $d_t(\boldsymbol{X})$, $t \in \mathcal{Y}$, each discriminating between $\boldsymbol{g}_t^f(\boldsymbol{X})$ and $\boldsymbol{g}_t^c(\boldsymbol{X})$. The training objective in equation (1) becomes

$$d_t(\cdot) = \underset{d(\cdot)}{\operatorname{argmin}} -P(Y=t)\mathbb{E}[\log d(\boldsymbol{g}_t^f(\boldsymbol{X}))|Y=t] - P(Y \neq t)\mathbb{E}[\log(1 - d(\boldsymbol{g}_0^c(\boldsymbol{X})))|Y \neq t]. \tag{24}$$

The goal of the factual and counterfactual generators is still to convince the discriminator that they are factual. The training objective in equation (2) becomes

$$\boldsymbol{g}_t^f(\cdot) = \underset{\boldsymbol{g}(\cdot)}{\operatorname{argmax}} \mathbb{E}[h^f(d_t(\boldsymbol{g}(\boldsymbol{X})))|Y=t], \quad \boldsymbol{g}_t^c(\cdot) = \underset{\boldsymbol{g}(\cdot)}{\operatorname{argmax}} \mathbb{E}[h^c(d_t(\boldsymbol{g}(\boldsymbol{X})))|Y \neq t]. \tag{25}$$

---

[3]There is no further differentiation of the counterfactual generators. In other words, the game is still played by three players. The counterfactual generator of each class, no matter what the ground truth label is, is considered as only one player.

Table 3: Statistics of the datasets used in this paper.

| Datasets | | Train | | Dev | | Annotation | |
|---|---|---|---|---|---|---|---|
| | | # Pos | # Neg | # Pos | # Neg | # Pos | # Neg |
| Amazon (single-aspect) | Book | 10,000 | 10,000 | 2,000 | 2,000 | 73 | 27 |
| | Electronic | 10,000 | 10,000 | 2,000 | 2,000 | 261 | 51 |
| Beer (multi-aspect) | Apperance | 16,890 | 16,890 | 6,627 | 2,103 | 923 | 13 |
| | Aroma | 15,169 | 15,169 | 6,578 | 2,218 | 848 | 29 |
| | Palate | 13,652 | 13,652 | 6,739 | 2,000 | 785 | 20 |
| Hotel (multi-aspect) | Location | 7,236 | 7,236 | 906 | 906 | 104 | 96 |
| | Service | 50,742 | 50,742 | 6,344 | 6,344 | 101 | 98 |
| | Cleanliness | 75,049 | 75,049 | 9,382 | 9,382 | 97 | 99 |

where $h^f(\cdot)$ and $h^c(\cdot)$ are monotocially increasing functions satisfying

$$xh^f\left(\frac{x}{x+a}\right) \text{ is convex in } x, \quad \text{and} \quad xh^c\left(\frac{a}{x+a}\right) \text{ is concave in } x, \quad \forall x, a \in [0, 1]. \tag{26}$$

Providing a theoretical guarantee for this multi-class case will be our future work.

## B  Additional Experiments and Details

### B.1  Datasets

The construction process of the three binary classification datasets we used is described below and some statistics of these datasets are summarized in table 3.

**Amazon review:** The original dataset contains customer reviews for 24 product categories. For each product domain, we filter the reviews that are with the patterns "pros: [...] cons: [...]". The goal is to select reviews that separately state the pros and cons of a product so that we could generate both factual and counterfactual rationales automatically using template matching. Specifically, we consider the first 100 tokens after the words "pros:" and "cons:" as the rationale annotations for the positive or negative prediction, respectively. After the filtering, only the domain of book and electronic have sufficient data for evaluation. Thus, we only include these two domains for our experiments. Since the data for both domains are notoriously large, we randomly select 10,000 examples with ratings of two as negative reviews and 10,000 reviews with ratings of four as positive ones. The reason to use ratings of two and four is that we hope to incorporate the data with both positive and negative aspects in a single review. The validation set contains another 2,000 examples for each rating.

**Beer review:** The beer subset introduced by Lei *et al.* [19] for rationalization contains reviews with ratings (in the scale of [0, 1]) from three aspects: appearance, aroma, and palate. Following the same evaluation protocol [8], we consider a classification setting by treating reviews with ratings $\leq 0.4$ as negative and $\geq 0.6$ as positive. Then we randomly select examples from the original training set to construct a balanced set. In addition, the dataset also includes sentence-level annotations for about 1,000 reviews. Each sentence is annotated with one or multiple aspects label, indicating which aspect this sentence belonging to. We use this set as factual evidence to automatically evaluate the precision of the extracted rationales.

**Hotel review:** The dataset contains reviews of the following three aspects: location, cleanliness, and service. We use the same data provided by Bao *et al.* [8], where reviews with ratings $> 3$ are labeled as positive and those with $< 3$ are labeled as negative. Similarly, for each aspect, the dataset contains 200 examples with human annotations, which explains why a particular rating is given.

### B.2  Experiment settings

The details of our experiment settings are as follows:

**Objective evaluation:** As we mentioned in the main paper, we compare the generated rationales with the human annotations and report the precision, recall and F1 score. For fair comparison, the evaluation is conditioned on a similar *actual* sparsity level in factual rationales (the target sparsity level is set to 10% for both beer and hotel review, and 20% for the Amazon dataset), which requires tuning the hyperparameters $\lambda_1$, $\lambda_2$ and $\alpha$. However, only the annotation set contains a fairly small number of the annotated rationales, it is hard to separate the set or generate more examples for the validation purpose. In order to control the actual sparsity and avoid overfitting the annotation set, we adopt the following setup to determine the hyperparameters. For each method, each hyperparameter is drawn from five candidate values, and we report the best test performance within $\pm 2\%$ of the preset sparsity level on factual rationales. This experiment setting is consistent with the one that has been reported previously [19].

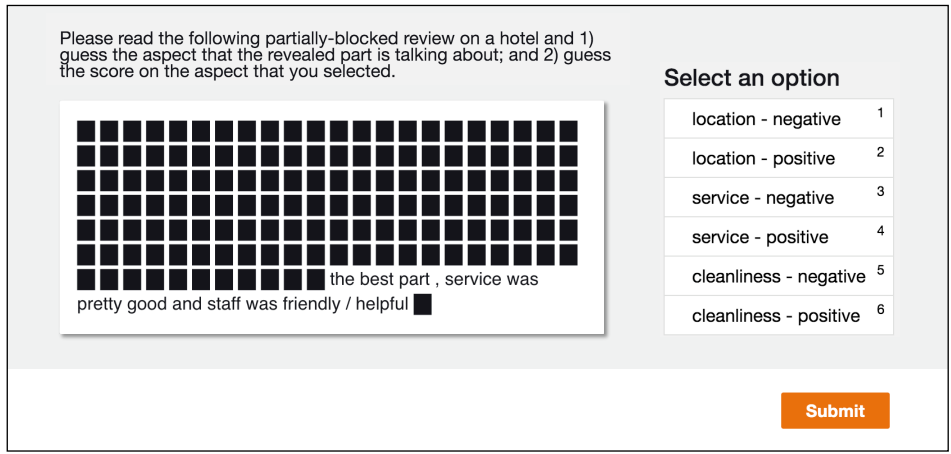

Figure 5: A screenshot of the user interface of the subjective evaluation on hotel review.

As for determining the number of training steps, that for RNP and the POST-EXP predictor is done by maximizing the sentiment prediction accuracy on the dev set[4], because their training goals are both to maximize the sentiment prediction accuracy. On the other hand, since the dev set does not have rationale annotations, the number of training steps of POST-EXP generators and our method has to be preset, guided by the principle to match what is needed for RNP to achieve a good dev performance, which is around 15 epochs. It is worth mentioning that all algorithms are trained on the training set, except for the POST-EXP generators, which are directly optimized based on the examples we evaluate. This is because the training of the generators does not require any supervision, as is the case for other gradient-based methods for model interpretations.

**Subjective evaluation:** Figure 5 shows the screenshot of an example test in the subjective evaluation on multi-aspect hotel review. As can be seen, the subjects are presented with text with blocks. Only the rationales are revealed. The subjects are asked to guess both the aspect and the sentiment of the aspect. Figure 6 shows the instructions that were presented to the subjects before the test. For the single-aspect datasets (Amazon review), the test is almost the same, except that the subjects do not need to guess the aspect.

### B.3 Additional results

Additional experiment results are shown below:

**Understanding the aroma aspect of beer reviews:** To better understand the lower performance of CAR on the aroma aspect of the beer reviews, we illustrate two examples of generated rationales in figure 7. We see that most of the factual and counterfactual rationales that are both only a subset of the ground truth annotation. This because the annotations contain a mix of sentiments. In the second example, we also see the counterfactual generator selects other text spans to prevent punished by the sparsity constraint.

**Additional illustrative results:** In addition to table 3 in the main paper, we include more illustrative results in table 8 and 9. Particularly, table 8 visualizes the generated rationales on the electronic domain of Amazon reviews while the latter table includes examples from the multi-aspect datasets.

---
[4]Although the dev set does not have rationale labels, it has sentiment labels.

**Instructions:**

We will be using a hotel review dataset, where reviewers are asked to write a review on each hotel and assign scores to the following aspects of the hotel: **location, service and cleanliness**. In each test, you will be seeing one such review, but the review is **partially blocked**. Each blocked word or punctuation will be shown as a "■". For example, the following sentence

*"This is good."*

with "is" and "." blocked will be shown as

*"This ■ good ■"*

The revealed part is supposed to talk about **only one of the three aspects**. Your task is to 1) guess what aspect the revealed part is talking about; and 2) guess the score that the reviewer assigned to the aspect that you guessed. There are only two possible scores:

- Negative;
- Positive.

For example, if you guess that the aspect is "service" and its score is "negative", please choose the option that says "service - negative".

**Please note that:**

- The reviews are on multiple aspects of the hotel, including location, service and cleanliness, but you are only asked to guess the score of the **aspect that you guess**.
- In some cases, the block will try to trick you into guessing a wrong score. For example

  *"This is not bad."*

  should have been a positive review, but the block may be like

  *"This is ■ bad."*

  **Please be ware of such tricks**.
- In some cases the review may be blocked so much that it is very hard to guess the aspect or score. However, you are still asked to guess the score to the best of your ability. Make random guess if necessary. Sometimes you will see the word "<unknown>". This is a representation of some rare words that are not in the vocabulary of our system. Feel free to ignore it.

Figure 6: Instructions of the subjective evaluation on hotel review.

Table 4: Subject evaluation results on the beer review dataset. 'Sentiment' and 'Aspect' are the marginal performances of the "Joint" results.

| Beer Review | | Appearance | | Aroma | | Palate | |
|---|---|---|---|---|---|---|---|
| | | Factual | Counter | Factual | Counter | Factual | Counter |
| Sentiment | RNP | 0.72 | - | 0.72 | - | 0.61 | - |
| | POST-EXP | **0.86** | 0.64 | 0.77 | **0.64** | 0.70 | 0.57 |
| | CAR | 0.85 | **0.68** | **0.81** | 0.59 | **0.85** | **0.69** |
| Aspect | RNP | 0.72 | - | **0.70** | - | **0.80** | - |
| | POST-EXP | 0.52 | 0.43 | 0.53 | **0.39** | 0.48 | 0.49 |
| | CAR | **0.72** | **0.58** | 0.48 | 0.17 | 0.75 | **0.51** |
| Joint | RNP | 0.54 | - | **0.57** | - | 0.50 | - |
| | POST-EXP | 0.46 | 0.28 | 0.44 | **0.24** | 0.35 | 0.25 |
| | CAR | **0.59** | **0.41** | 0.42 | 0.13 | **0.68** | **0.43** |

Similarly, we observe that CAR produces meaningful justifications for both factual and counterfactual labels.

**Additional subjective evaluations:** Table 4 and 5 illustrate comprehensive results of the subjective evaluations for the beer and hotel reviews, respectively. The results in the bottom section (*i.e.,* "joint") of the tables are included in figure 4 in the main paper. Recall that in the multi-aspect subjective experiments, the subjects need to answer both aspect category and the sentiment valence. Both tables include the accuracy of each marginal statistics of the results, which are shown as "sentiment" and "aspect" in the tables. The former one records sentiment accuracy regardless of whether the aspect is correctly classified while the latter one represents the aspect accuracy only.

| *Beer - Aroma* | Label - Positive |
|---|---|

poured from a bottle into a pint glass . a : pours a ruby amber color with 1/4 ' thin head and some lacing . this is a good looking **beer . s : floral hops and citrus burst into the nostrils . some slight malt and pine aromas also present** . t : sweet citrus flavors dominate the initial sip followed by a significant hop presence in the middle to finish . some slightly unappealing bitter aftertaste in the finish . m : this beer has the feel of a good spring beer . it is light and thin on the tongue with the presence of prickly carbonation . overall this is a good beer . it is very similar in flavor to nugget nectar , but the flavors and character are not as prominent as in its imperial amber ale cousin . aside from the slightly bitter aftertaste , the beer is well balanced and one of the tastier amber ales that i have had .

| *Beer - Aroma* | Label - Positive |
|---|---|

small somewhat creamy light brown head that mostly diminished . **pitch black color . aroma of bread , molasses , caramel , chocolate , coffee** , toasted **malt . light alcoholic sniff** , plum , prune , date , licorice . very complex aroma . full bodied with an alcoholic texture and a soft carbonation . excellent balance between **the moderated sweet malt** and bitter flavor . berries sweet and lingering finish . smooth and very nice thick beer . definitely one of my favorite ris .

Figure 7: Examples of CAR generated rationales on the aroma aspect of the beer reviews. The provided ground truth annotations are shown in the blue text. All rationale words are **bold and underlined**. To distinguish the factual and counterfactual generation, we use blue highlight to represent the former one while red highlight for the latter one.

| *Amazon - Electronic* | Label - Positive |
|---|---|

**these are good , solid headphones . i ca n't say they blew me away , <unknown> , the frequency response is very flat** . with & # 34 ; street by 50 cent & # 34 ; on the box , i was kind of expecting them to be seriously bass-heavy , but that 's not the case . **i listened to quite a variety of different music , and was pleased with their performance across the board.i hesitate to refer** to & # 34 ; noise canceling , & # 34 ; since there is no active nc going on here , but the ear cups do a decent job of blocking out outside sounds . they are soft and have smooth covers , and i can go about an hour **before they start to get uncomfortable . not bad** , but could be better.the star wars themed design and accessories are pretty cool . this <unknown> fett model will only be recognizable to the hardcore fan , but the design is nice looking anyway .

| *Amazon - Electronic* | Label - Negative |
|---|---|

**i purchased the edimax <unknown> to take advantage of** the 5ghz band from my asus dark knight router , **which is an awesome** router by the way . **installation from the cd was easy** and then updating the firmware using a download from the edimax website was **also easy . however , i was really disappointed in the performance .** i used inssider 3 to verify connection to the 2.4 or 5 ghz band . 95 % of the time the edimax would connect to the 2.4ghz band and not the 5ghz band even though the 5ghz band was showing to be present via inssider <unknown> when it was connected to the 5ghz band the best speed was <unknown> download , which is **not too bad** . yet most of the time , the edimax connected to the 2.4ghz with download speeds of <unknown> or less . the strangest thing i noticed about the edimax adapter was that is did not sustain a consistent signal from the router only 8 feet away . during the ookla download tests , the signal always had a saw tooth appearance and **the speed needle swung <unknown> 'm going to put the tp** link <unknown> **back on the laptop which consistently beats** the edimax adapter . the tp link download speeds were 38 to 55mbs on the 2.4ghz <unknown> to put these numbers into perspective , i have fios <unknown> and get wired download speeds of about <unknown> and wired uploads of <unknown> .

Figure 8: Examples of CAR generated rationales on the electronic domain of the Amazon reviews. All selected words are **bold and underlined**. Factual generation uses blue highlight while the counterfactual uses red one.

Table 5: Subject evaluation results on the hotel review dataset. "Sentiment" and "Aspect" are the marginal performances of the "Joint" results.

| Hotel Review | | Location | | Service | | Cleanliness | |
|---|---|---|---|---|---|---|---|
| | | Factual | Counter | Factual | Counter | Factual | Counter |
| Sentiment | RNP | 0.78 | - | 0.87 | - | 0.88 | - |
| | POST-EXP | **0.84** | 0.53 | 0.89 | 0.55 | 0.84 | 0.39 |
| | CAR | 0.83 | **0.53** | **0.93** | **0.56** | **0.96** | **0.62** |
| Aspect | RNP | 0.83 | - | 0.81 | - | 0.42 | - |
| | POST-EXP | 0.65 | 0.62 | 0.81 | 0.58 | 0.27 | **0.15** |
| | CAR | **0.87** | **0.77** | **0.84** | **0.62** | **0.52** | 0.13 |
| Joint | RNP | 0.66 | - | 0.71 | - | 0.40 | - |
| | POST-EXP | 0.55 | 0.34 | 0.76 | 0.35 | 0.24 | 0.06 |
| | CAR | **0.74** | **0.41** | **0.78** | **0.36** | **0.52** | **0.08** |

*Beer - Appearance*      Label - Positive

poured into pint glass . a : used motor **oil color . two inches of frothy light brown head with excellent retention and quite a bit** of lacing . nice cascade going for a while . s : oatmeal is the biggest component of the aroma . not any hops content . a bit fusely and a bit of alcohol . t : tastes like slightly sour nothing . i do n't know **what the** hell made this dark because their is no crystal malt or roasted barley component in the taste . this sucks . m : **light body , really light body like water . carbonation** is fine , but that 's about it . d : this is slightly sour water . it fucking sucks . how the hell does anybody like this ?

*Beer - Appearance*      Label - Negative

got this at de bierkoning in amsterdam . from a bottle into a mug . appearance : pours **a very small and thin off-white head that quickly disappears** . lower level carbonation evident , which quickly ceases as well . colour is more brown than red , like a sienna with some faint hints of rust on the sides . **zero film stays and** no lacing . relatively lack **luster** , even for the style . smell : a medium strong nose of sweet caramel malt , a bit of toffee and maybe some red fruit in the back . a little one dimensional , but nice enough . taste : sweet caramel malt in the front with some toffee as well , and finishes with a burst of spicy alcohol at the end . aftertaste **is nice , mild and long lasting** , consisting of some spicy alcohol and a slight touch of bitterness . not bad ; more impressive than the nose . palate : medium body and medium carbonation . **relatively creamy on** the palate , goes down quite smooth with a small alcohol bite at the very end . finishes on the stickier side of the spectrum . overall : not bad , but not memorable . the look was certainly disappointing , but it was pretty average other than that . drinkable enough , but not worth seeking out .

*Hotel - Service*      Label - Positive

i would definitely recommend this hotel to anyone who has plans at the staples center or nokia theater **. the staff was friendly and helpful , the rooms were clean , and there was plenty of entertainment and dining options within walking distance . they provide robes , slippers , and** wonderful toiletries ! the shower was beautiful but **the head was not working properly and we requested a non smoking room but there were n't any available . the smell of stinky cigarettes really disgusted me** but if they had a non smoking room available i honestly would not have any complaints . i recommend taking a car to and from the airport . i would stay at this hotel again for sure .

*Hotel - Service*      Label - Negative

**i loved the hotel and the hotel room . my daughter and i were very impressed .** we had a lovely room and for being a hilton honors **member we even received a free fruit basket** and two bottles of water . please beware however and always check your bank account for any additional charges . we had used my daughters credit card to hold the room but asked for my visa debit card to be charged for the final amount . we were both charged for the room **and it took numerous phone calls to receive a credit . we were also charged for several additional charges that we** should not have been . the billing department was extremely rude as we called them numerous times to get the matter straightened out . as for the hotel i would **highly recommend it and also for it 's wonderful** location and shuttle . i just did n't appreciate the cutomer service afterwards and the fact that it took me so long to get the matter taken care of . just be very careful looking over your final billing .

Figure 9: Examples of CAR generated rationales on the multi-aspect datasets. All selected words are **bold and underlined**. Factual generation uses blue highlight while the counterfactual uses red one.